\documentclass{ecai}
\usepackage{times}
\usepackage{graphicx}
\usepackage{latexsym}
\usepackage[algoruled,linesnumbered]{algorithm2e}
\usepackage{caption}
\usepackage{amsmath}
\usepackage{nameref}
\usepackage{varioref}
\usepackage{hyperref}
\usepackage{cleveref}
\usepackage{float}
\usepackage{url}
\DeclareMathOperator*{\argmax}{arg\,max}

\usepackage{footnotebackref}
\usepackage{blindtext}
\ecaisubmission   % inserts page numbers. Use only for submission of paper.
\usepackage[bottom,flushmargin,hang,multiple,para]{footmisc}

\newcommand{\method}{SCDV-MS}

\begin{document}

\title{Improving Document Classification with Multi-Sense Embeddings}

\author{Vivek Gupta\institute{University of Utah, email: vgupta@cs.utah.edu} \and Ankit Kumar\institute{IIT Karaggpur, email: ankit.kgpian@gmail.com} \and Pegah Nokhiz\institute{University of Utah, email: pnokhiz@cs.utah.edu} \and Harshit Gupta\institute{IIT Guwahati, email: harshitgupta@iitg.ac.in} \and Partha Talukdar\institute{IISC Bangalore, email: ppt@iisc.ac.in }}

\maketitle
\bibliographystyle{ecai}

\begin{abstract}
Efficient representation of text documents is an important building block in many NLP tasks. Research on long text categorization has shown that simple weighted averaging of word vectors for sentence representation often outperforms more sophisticated neural models. Recently proposed Sparse Composite Document Vector (SCDV) \cite{mekala2017scdv} extends this approach from sentences to documents using soft clustering over word vectors. However, SCDV disregards the multi-sense nature of words, and it also suffers from the curse of higher dimensionality. In this work, we address these shortcomings and propose SCDV-MS. SCDV-MS utilizes multi-sense word embeddings and learns a lower dimensional manifold. Through extensive experiments on multiple real-world datasets, we show that SCDV-MS embeddings outperform previous state-of-the-art embeddings on multi-class and multi-label text categorization tasks. Furthermore, SCDV-MS embeddings are more efficient than SCDV in terms of time and space complexity on textual classification tasks. We have released SCDV-MS source code with the paper. \footnote{\url{https://github.com/vgupta123/SCDV-MS}}
\end{abstract}

\section{Introduction}
\label{sec:introduction}

%Word Embedding%
Distributed word embeddings such as word2vec \cite{mikolov2013linguistic} are effective in capturing the semantic meanings of the words by representing them in a lower-dimensional continuous space. A smooth inverse frequency-based word vector averaging technique for sentence embeddings was proposed by \cite{arora2016simple}. However, because the final representation is in the same space as the word vectors, these methods are only capable of capturing the meaning of a single sentence. Thus, embedding a large text document in a dense, low-dimensional space is a challenging task.

%Document Embedding%
\cite{mekala2017scdv} attempted to resolve this problem by proposing a clustering-based word vector averaging approach (SCDV) for embedding larger text documents. SCDV embeds each document by cluster-based averaging, thus representing each document in a more representative space than the original vectors. This model combines the word embedding models with a latent topic model where the topic space is learned efficiently using a soft clustering technique over embeddings. The final document vectors are also made sparse to reduce time and space complexities in several downstream tasks. 

SCDV has many shortcomings: applying thresholding-based sparsity on the document representations can be unreliable since it is highly sensitive to the number of documents in the corpus. SCDV does not utilize the word contexts to disambiguate word sense for learning sense-aware document representations. Ignoring the multi-sense nature of the words during the representation leads to cluster ambiguity.  SCDV neglects the negative additive effect of common words such as \textit{`and', `the',} etc. during final document representations. Lastly, the documents represented by SCDV suffer from the curse of high dimensionality and cannot be utilized for deep learning applications which require low-dimensional continuous representations. To overcome these challenges, we proposed a novel document representation technique, namely \method{}. Our proposed \method{} mitigated the above shortcomings by the following contributions.

% However, applying thresholding-based sparsity on the document representations can be unreliable since it is highly sensitive to the number of documents in the corpus. SCDV also ignores the multi-sense nature of the words during the representation. Additionally, SCDV does not utilize the word contexts to disambiguate word sense for learning better representations. Moreover, SCDV neglects the negative additive effect of common words such as \textit{`and', `the',} etc. during the final document representations. Lastly, the documents represented by SCDV suffer from the curse of high dimensionality and cannot be utilized for deep learning applications. To overcome the above challenges, we proposed a new document representation technique, namely \method{}. \method{} improved the SCDV representations by the following contributions.

\begin{enumerate}
\setlength\itemsep{0.2em}
\item To overcome the problem of cluster ambiguity \method{} replaced the single sense word vector representations with multi-sense context-sensitive word representations to resolve word sense disambiguation.
% \vspace{-0.2em}
\item \method{} removed the noise in the final representation by applying a threshold-based sparsity directly on fuzzy word cluster assignments instead of the document representations. Sparser representations result in better performance, lower time and space complexities.
% \vspace{-0.2em}
\item To overcome the noisy negative additive effect of common words such as \textit{`and', `the',} etc. \method{} learned and used Doc2VecC-initialized \cite{chen2017efficient} robust word vectors to zero out common and high frequent words.
% \vspace{-0.3em}
\item Lastly, we showed that the sparse word-topic vectors can be projected into a non-linear local neighborhood preserving a manifold to learn continuous distributed representations much more effectively and efficiently than SCDV which proves to be useful for deep learning application.
\end{enumerate}

Overall, we show that: disambiguating the multiple senses of words based on their context words (adjacent words) can lead to better document representations. Sparsity in representations is helpful for effective and efficient lower-dimensional manifold representation learning. Representation noise at words' level has a significant impact on the final downstream tasks. 

\vspace{0.3em}
\noindent In section \ref{sec:introduction}, we provided a brief introduction to the problem statement. In section \ref{sec:relatedwork} we discuss the related work in document representations. In section \ref{sec:proposeAlgo}, we describe the proposed algorithm and then discuss the proposed modifications compared to SCDV in section \ref{sec:proposemodification}. We move on to experiments in section \ref{sec:experiments}, followed by conclusions in section \ref{sec:conclusions}. 

\section{Related Work}
\label{sec:relatedwork}

For document representation, averaging of word-vectors with an unweighted scheme was proposed by \cite{smolensky1990tensor,mitchell-lapata-2008-vector,mitchell2010composition,mikolov2013distributed}. \cite{pranjal2015weighted} extended the previous simple averaging model by tf-idf weighting of word vectors to form document vectors. \cite{le2014distributed} proposed paragraph models (PV-DM and PV-DBOW) similar to word vector models (CBoW and SGNS) by treating each paragraph as a pseudoword. \cite{socher2013recursive} used a parse tree to train a Recursive Neural Network (RNN) with supervision.  In addition, several neural network models such as seq2seq models: Recurrent Neural Networks (RNN) \cite{mikolov2010recurrent}, Long Short Term Memory Networks (LSTM) \cite{gers2002learning} and a hierarchical model: Convolutional Neural Networks (CNN) \cite{kim2014convolutional,kalchbrenner2014convolutional} were proposed to capture syntax while representing documents. \cite{wieting2015paraphrase} used supervised learning over the Paraphrase Dataset (PPDB) to learn Paraphrastic Sentence embeddings (PSL). Later, \cite{wieting2015towards} also propose an optimization of word embeddings based on a neural network and a cosine similarity measure.

Several models such as WTM \cite{wtm}, TWE  \cite{liu2015learning}, NTSG \cite{liu2015learning}, LTSG \cite{ltsg},  w2v-LDA \cite{wtvlda}, TV+MeanWV \cite{tvMeanWV}, Topic2Vec \cite{topic2vec}, Gaussian-LDA \cite{gaussianlda}, Lda2vec \cite{lda2vec}, ETM \cite{dieng2019topic}, D-ETM \cite{dieng2019dynamic} and MvTM \cite{mvtm} combine topic modeling \cite{Blei:2003} with word vectors to learn better word and sentence representations. \cite{kiros2015} cast the distributional hypothesis to a sentence level by proposing skip-thought document vectors.  Recently, two deep contextual word embeddings namely ELMo \cite{Peters:2018} and BERT \cite{devlin2018bert} were proposed. These contextual embeddings perform as state of the art on multiple NLP tasks since they are very effective in capturing the surrounding context. Interestingly, \cite{li2015multi} checks the effect of using multi-sense embeddings on various NLP tasks. However, our goal is different and aim at effectively using multi-sense words embeddings to learn better document representations. A hard clustering-based averaging of word vectors was proposed by \cite{vivek,hadi2019vector} to form document vectors. \cite{gupta2019unsupervised} extended the approach with a better partitioning technique and tried it on other natural language tasks. \cite{mekala2017scdv} further improved the state-of-the-art SCDV by using fuzzy clustering and tf-idf weighted word averaging. Their method outperformed earlier models on several NLP tasks.

\section{Proposed Algorithm \method{}}
\label{sec:proposeAlgo}

In this section, we will describe our new proposed algorithm \ref{algo:DV} in details. The algorithm is similar to SCDV \cite{mekala2017scdv}, but with important modifications and has three main components as described below:

\begin{algorithm}[h!]
    \SetAlgoNoLine
    \setlength{\textfloatsep}{0.0pt}
    \KwData{Documents D$_{n}$, $n = 1, \ldots ,N$}
    \KwResult{Document vectors $\vec{dv}_{D_{n}}$, $ n = 1, \ldots ,N$}
    \tcc{Word Sense Disambiguation}
    \vspace{0.5em}
    Use adagram for word sense disambiguation\;
    \label{algo:sensedisambiguatestart}
    Annotate each word with a sense according to the neighboring context words\;
    Obtain word vectors ($\vec{wv}_i$) on annotated corpus using Doc2VecC\;
    \For {each word $w_i$ $\in V$}{
    obtain idf values, idf(w$_{i}$),\,$i = 1..|V|$ \;
    \tcc{$|V|$ is the vocabulary} 
    }
    \label{algo:sensedisambiguateend}
    \label{algo:gmmstart}
    \vspace{0.5em}
    \tcc{Word Vector Clustering}
    \vspace{0.5em}
    Fuzzy clusters $\vec{wv}$ in $K$ clusters\;
    Each word $w_{i}$ and cluster $c_{k}$, obtain P$(c_{k}|w_{i})$ \; 
    $\vec{\text{SP}(c|w_{i})}$ = make-sparse($\vec{\text{P}(c|w_{i})}$)\;
    \label{algo:gmmend}
    \label{algo:wtvstart}
    \vspace{0.5em}
    \tcc{Word Topic Vectors}
    \vspace{0.5em}
    \For{each word $w_i$ $\in$ $V$}{
        \For{each cluster $c_k$}{
             $\vec{wcv}_{ik}$ $=$ $\vec{wv_i}$ $\times$ SP$(c_{k}|w_{i})$\;
        }
        $\vec{wtv}_{i}$ $=$ idf(w$_i$) $\times$ $\bigoplus_{k=1}^K$ $\vec{wcv}_{ik}$ \;
        \tcc{$\bigoplus$ is concatenation}
        \tcc{Optional: Manifold Learning}
        $\vec{rwtv}_{w}$ = manifold-proj($\vec{wtv}_{w}$)\;
    }
    \vspace{0.5em}
    \tcc{SCDV-MS Representation}
    \vspace{0.5em}
    \label{algo:wtvend}
    \For{ $n\in (1..N)$}{
        \tcc{initialize vectors}
        $\vec{dv}_{D_n}$ = $\vec{0}$\;
        \For{word $w_i$ in \text{D}$_n$}{
            $\vec{dv}_{D_n}$ += $\vec{wtv}_{w_i}$\;}}
\caption{SCDV-MS}
\label{algo:DV}
\end{algorithm}
\subsection{Word Sense Disambiguation (Algo \ref{algo:DV}: \ref{algo:sensedisambiguatestart} - \ref{algo:sensedisambiguateend}):}
\label{algo:wsd}
We employed the widely-used AdaGram \cite{bartunov2016breaking} algorithm to disambiguate the multi-sense words in our corpora. We chose AdaGram because it's a nonparametric Bayesian extension of Skip-gram which automatically learns the counts of the senses of the multi-sense words and their representations. \footnote{One could also replace AdaGram with \cite{dai2017mixture} and \cite{athiwaratkun2018probabilistic}} We first trained the AdaGram algorithm on the training corpora. \footnote{ \url{https://github.com/sbos/AdaGram.jl}} We used the trained model to annotate the words with the corresponding word senses in all train-test examples. We then trained the Doc2vecC algorithm on an annotated corpus to obtain the final multi-sense word vectors. Lastly, we obtained the idf values of words of the vocabulary which we will use as a means for weighting the rare words (Lines 4-6 Algo \ref{algo:DV}).

\subsection{Word Vector Clustering (Algo \ref{algo:DV}: \ref{algo:gmmstart} - \ref{algo:gmmend})}
\label{algo:wvc}
Similar to the SCDV approach, we clustered the word embeddings using Gaussian Mixture Models (GMM), which is a soft clustering technique, and obtained the word-cluster assignments probabilities P$(c_{k}|w_{i})$.  Additionally, we made use of the fact that GMMs have an irrelevant noisy tail and made the cluster probability assignment $\vec{\text{P}(c|w_{i})}$ manually sparse by zeroing the values of P$(c_{k}|w_{i})$. Retaining only the top $l$ maximum P$(c_{k}|w_{i})$ and zeroing the rest $K-l$ values results in a sparse word-cluster assignment vector $\vec{\text{SP}(c|w_{i})}$ for each word. Here, $K$ represents the total number of clusters and $l$ is the sparsity constant ($l<<K$). One can use different values of $l$ for each word ($w_i$) depending on the values of P$(c_{k}|w_{i})$ . However, in our experiments we did not observe significant performance difference when $l$ is varied with respect to the words. 
\begin{equation}
\small
\text{SP}(c_k|w_{i}) = 
\begin{cases}
\text{P}(c_{k}|w_{i}) & \text{if } k \in \{k | \argmax_k^l{{\text{P}(c_{k}|w_{i})}}\} \\ 0 & \text{otherwise}
\end{cases}
\end{equation}
$ \argmax_k^l{{\text{P}(c_{k}|w_{i})}}$ outputs indices of the top $l$ maximum assignments.
\subsection{Word Topic Vectors (Algo \ref{algo:DV}: \ref{algo:wtvstart} - \ref{algo:wtvend})}
\label{algo:wtv}
Similar to SCDV, for each word $w_i$ $\in$ $V$, we created $K$ different word-cluster vectors of $d$ dimensions ($\vec{wcv}_{ik}$) by weighting the word's embedding with sparse probability distribution for the $k^{th}$ cluster,  i.e., $SP(c_{k}|w_{i})$. Next, we concatenated the word-cluster vectors  ($\vec{wcv}_{ik}$) which are $K$ word-topic vectors in total, into a $K \times d$ embedding vector. We then weighted it with inverse document frequency (idf) of $w_i$ to obtain word-topic vectors ($\vec{wtv}_i$). For all words appearing in a given document $D_n$, we computed the average of the corresponding projected lower dimensional word-topic vectors  $\vec{wtv}_i$ to obtain the document vector $\vec{dv}_{D_n}$. Furthermore, one can optionally project the ($\vec{wtv}_i$) into a lower dimensional continuous representation called reduced word topic vectors, ($\vec{rwtv}_i$), using manifold learning algorithms, namely Random Projection \cite{achlioptas2003database}, PCA \cite{abdi2010principal} and Denoising Autoencoders \cite{vincent2010stacked}. We can then use them instead of ($\vec{wtv}_i$) for document representation. We call this reduction-based representation method as R-SCDV-MS. Refer to section \ref{subsec:multisense} on manifold learning for more details.
\begin{equation}
% \vspace{-0.1cm}
\small
\vec{wcv}_{ik} = \vec{wv}_i \times \text{SP}(c_{k}|w_{i})
\end{equation}
\begin{equation}
\small
\vec{wtv}_{i} = \text{idf}(w_i) \times \oplus_{k=1}^K \vec{wcv}_{ik}
\end{equation}
\begin{equation}
\small
\vec{rwtv}_{i} = \text{manifold-proj}(\vec{wtv}_{i}) 
\end{equation}

$\bigoplus$ is the concatenation and manifold-proj is the manifold learning algorithm we utilized. Figures \ref{figure:pipeline1} and \ref{figure:pipeline2} show the flow-chart of high level flow of our proposed \method{} embedding.

% % Refer to Figure \ref{figure:illustration} for the high level flow of our proposed approach.
% \begin{figure*}[t]
% \centering
% \includegraphics[scale=0.22]{wordtopicvector.png}
% \vspace{-0.1em}
% \caption{Flowchart representing modified word topic vector computation.}
% \label{figure:pipeline1}
% \end{figure*}
% \begin{figure*}[t]
% \centering
% \includegraphics[scale=0.22]{documentvector.png}
% \vspace{-0.1em}
% \caption{Flowchart representing final document vector computation.}
% \end{figure*}

\section{Discussion on Proposed Modifications}
\label{sec:proposemodification} 

In this section, we will describe the modifications applied to the SCDV embeddings in details.
\subsection{Word Representation: Single Sense vs Multi Sense}
\label{subsec:multisense}

\method{} used a multi-sense approach instead of single sense word embeddings because SCDV does not disambiguate the senses of the words based on the context words used in the documents. \method{} performed an automatic word sense disambiguation using multi-sense word embeddings according to the context determined by the neighboring words to resolve cluster ambiguity for polysemous words. Table \ref{table:sensediasm} shows examples of multi-sense words along with their fitting context and the prominent words of the assigned clusters. 

\begin{figure}[!htbp]
\centering
\includegraphics[width=0.34\textwidth]{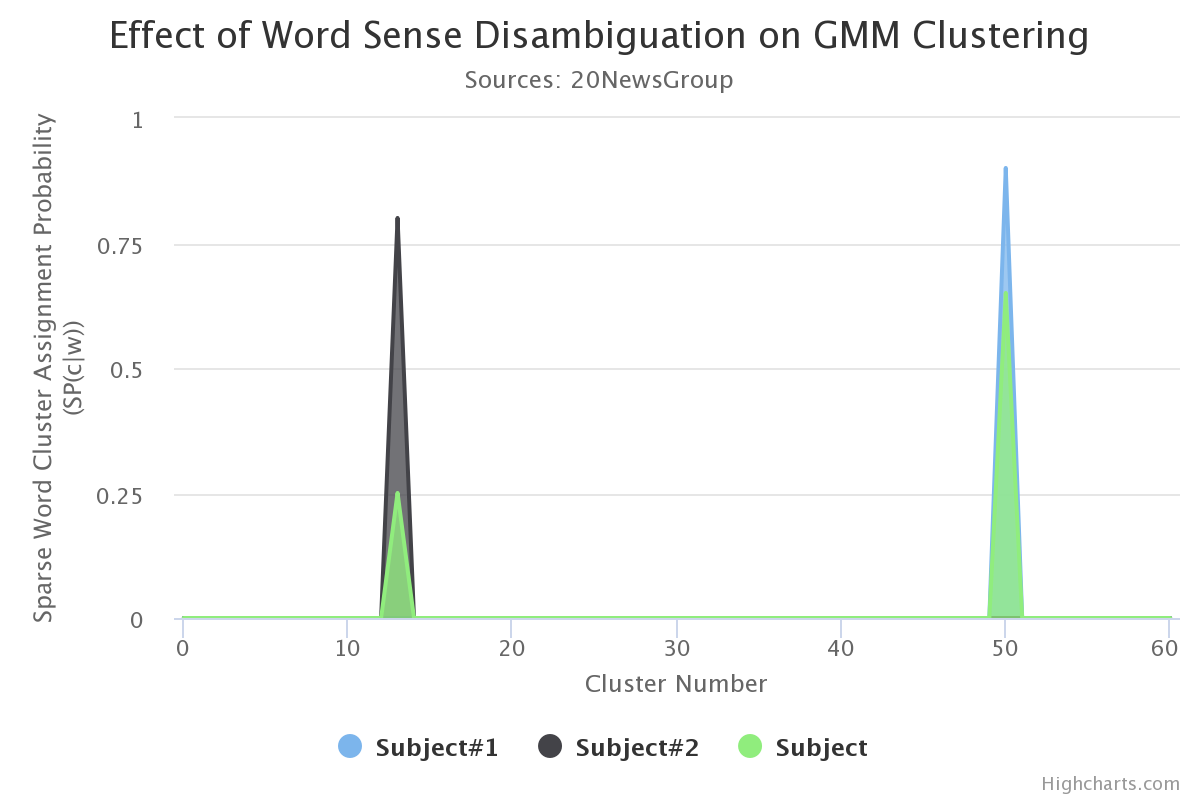}
\vspace{-0.5em}
\caption{Effect of sense disambiguation on word cluster assignment probabilities}
\label{figure:sence-disamiguation}
\vspace{-1.5em}
\end{figure}

Figure \ref{figure:sence-disamiguation} shows the effect of sense disambiguation on fuzzy GMM clustering on the 20NewsGroup dataset. The same word is assigned to different clusters depending on its context which helps in resolving the word cluster ambiguities, e.g., without sense disambiguation, the word \textit{'Subject'} belongs to cluster $13$ with probability $0.25$ and cluster $50$ with probability $0.65$. But after sense disambiguation we acquire two word embeddings of the word \textit{`Subject'}, i.e., \textit{`Subject\#1'} and \textit{`Subject\#2'}. \textit{`Subject\#1'} belongs to cluster $50$ with probability of $0.9$ and \textit{`Subject\#2'} belongs to cluster $13$ with probability $0.8$. So depending on the context in which word \textit{`Subject'} is used, the algorithm assigns \textit{`Subject'} to a single cluster based on its sense; thus word cluster ambiguity is resolved. We observe similar disambiguation effects for other polysemous words in the corpus.

\begin{table*}[h]
\captionsetup{font=small, skip=0pt}
\begin{center}
\caption{Examples of multi-sense words along with their context words and the corresponding prominent cluster words}
\vspace{0.5 em}
\small
\label{table:sensediasm}
\begin{tabular}{c|c|c} 
 \hline
  {\bf Multi-Sense Words } & {\bf Sentence (Context Words) } & {\bf  Prominent Cluster Words} \\
  \hline
  Subject & The math \textbf{subject} is a nightmare for many students & physics, chemistry, math, science \\
   & In anxiety, he sent an email without a \textbf{subject}  & mail, letter, email, gmail \\
  \hline
  & After promotion, he went to Maldives for spring \textbf{break}  & vacation, holiday, trip,  spring \\
  Break & \textbf{Breaking} government websites is common for hackers & encryption, cipher, security, privacy\\
   & Use \textbf{break} to stop growing recursion loops & if, elseif, endif, loop, continue \\
  \hline
  Unit & The S.I. \textbf{unit} of distance is meter & calculation, distance, mass, length \\
  & Multimeter shows a \textbf{unit} of 5V & electronics, KWH, digital, signal \\
  \hline
  Interest & His \textbf{interest} lies in astrophysics & information, enthusiasm, question\\
   & Bank’s \textbf{interest} rates are controlled by RBI & bank, market, finance,  investment \\
  \hline
\end{tabular}
\end{center}
\vspace{-1.8em}
\end{table*}

\subsection{Thresholding Word Cluster Assignments}
\label{subsec:hardthres}

In SCDV, the thresholding is applied to the final document vectors. However, applying the hard thresholding in an earlier stage in word cluster assignments ($P(c_i|w)$) results in better heavy tail noise removal and yields more robust representations. Thus, \method{} applied the hard thresholding directly on the word cluster assignments instead of the final document representations. Furthermore, applying sparsity over vocabulary words with fewer dimensions ($O(VK)$) instead of millions of documents ($O(NKd)$) results in higher efficiency ($Nd >> V$). Here, $N$ is the number documents, $V$ is vocabulary, $K$ is number of clusters, and $d$ is word vector dimensions. Empirically, on 20NewGroups we observe that about $98\%$ of entries in word-cluster assignments $(P(c_i|w))$  for all words are close to $0$ ($<$ $0.01$). For each word on average, the probability of cluster assignment ($P(c_i|w)$) for $58$ cluster assignments out of $60$ (variance of $1.56$) is less than $0.01$. Thus, applying thresholding at word-cluster assignment level is reasonable. 

\subsection{Doc2VecC vs SGNS Representations}
\label{subsec:doc2vecC}
In the SCDV approach, the SGNS  word vectors represent the common words (mainly stop words) as non-zero vectors. This makes the clusters redundant and generates a heavy tail noise. \method{} addressed this issue by using Doc2VecC \cite{chen2017efficient} which introduces corruption while doing context addition in word embeddings to help in learning robust word vector representations. In this approach, the common words of the corpus are forcefully learned as zeroed vectors. We observed that using Doc2VecC trained word vectors results in non-redundant diverse clusters. Thus, using Doc2VecC trained word vectors not only improves the performance but also reduces the feature size. There is no running time overhead for Doc2VecC compared to the SGNS.

\subsection{Low Dimensional Manifold Learning}
\label{subsec:manifold}
SCDV \cite{mekala2017scdv} represents documents as high dimensional sparse vectors. The SCDV approach showed that such vectors are useful for linear classification models. However, being useful for many downstream applications (especially the ones using deep learning models) requires a continuous low dimensional document representation similar to word vectors. To overcome this issue, \method{} projected the sparse word-topic vectors into a lower-dimensional manifold which preserves the local neighborhood using simple techniques such as random projection. Furthermore, the manifold learning is applied over word vocabularies instead of millions of documents, which is more efficient. Dimensionality reduction for SCDV-MS is roughly ($O(\frac{N}{V})$) (where $N$ is the number of documents and $V$ is the size of vocabulary) faster than SCDV.

\begin{figure*}[!htb]
\minipage{0.49\textwidth}
\captionsetup{font=small, skip=0.0pt}
  \includegraphics[width=\linewidth]{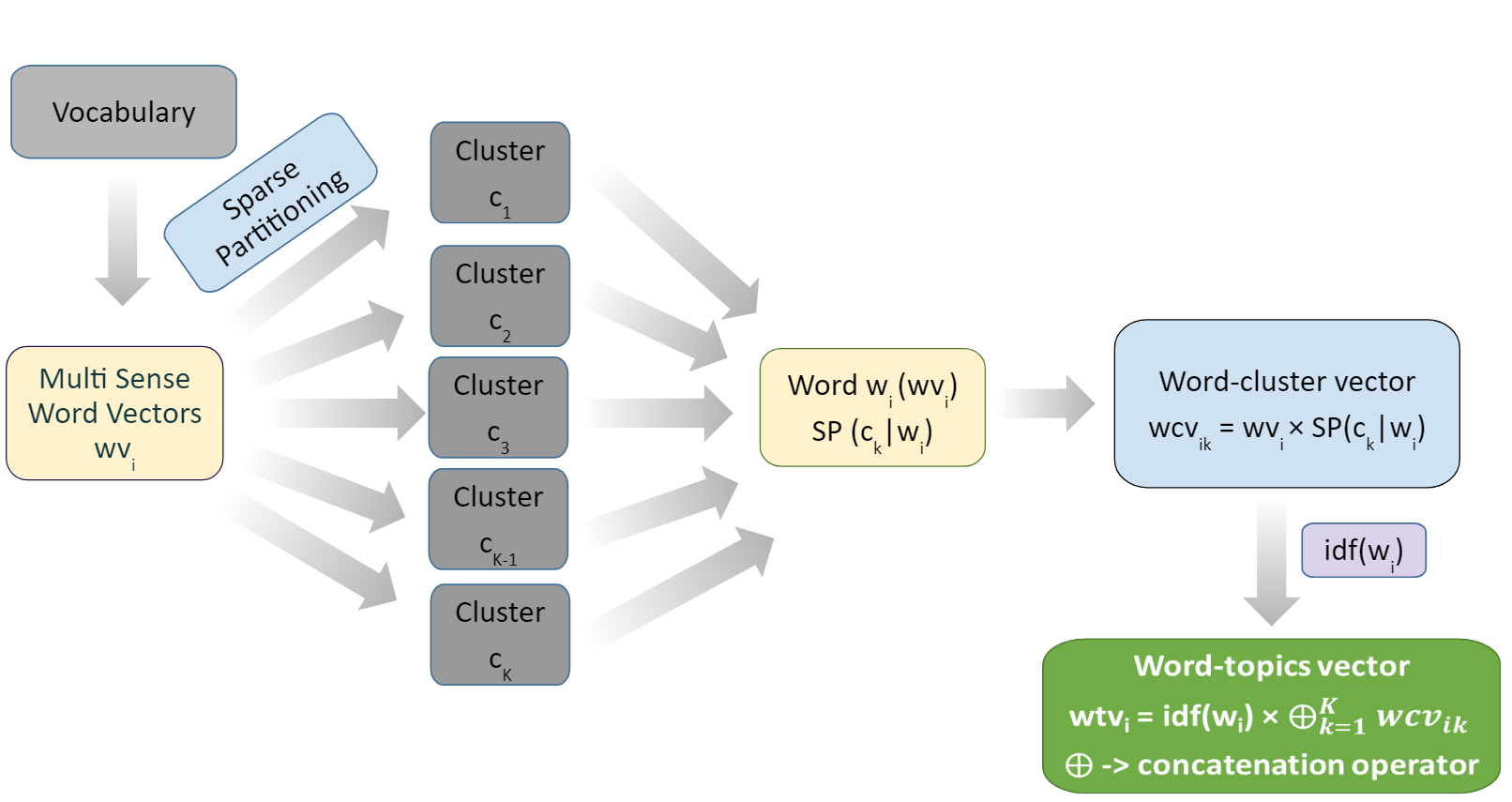}
  \vspace{0.5 em}
  \caption{Flowchart representing modified $\vec{wtv}$ computation.}\label{figure:pipeline1}
\endminipage
\hfill
\minipage{0.49\textwidth}%
\captionsetup{font=small, skip=0pt}
  \includegraphics[width=\linewidth]{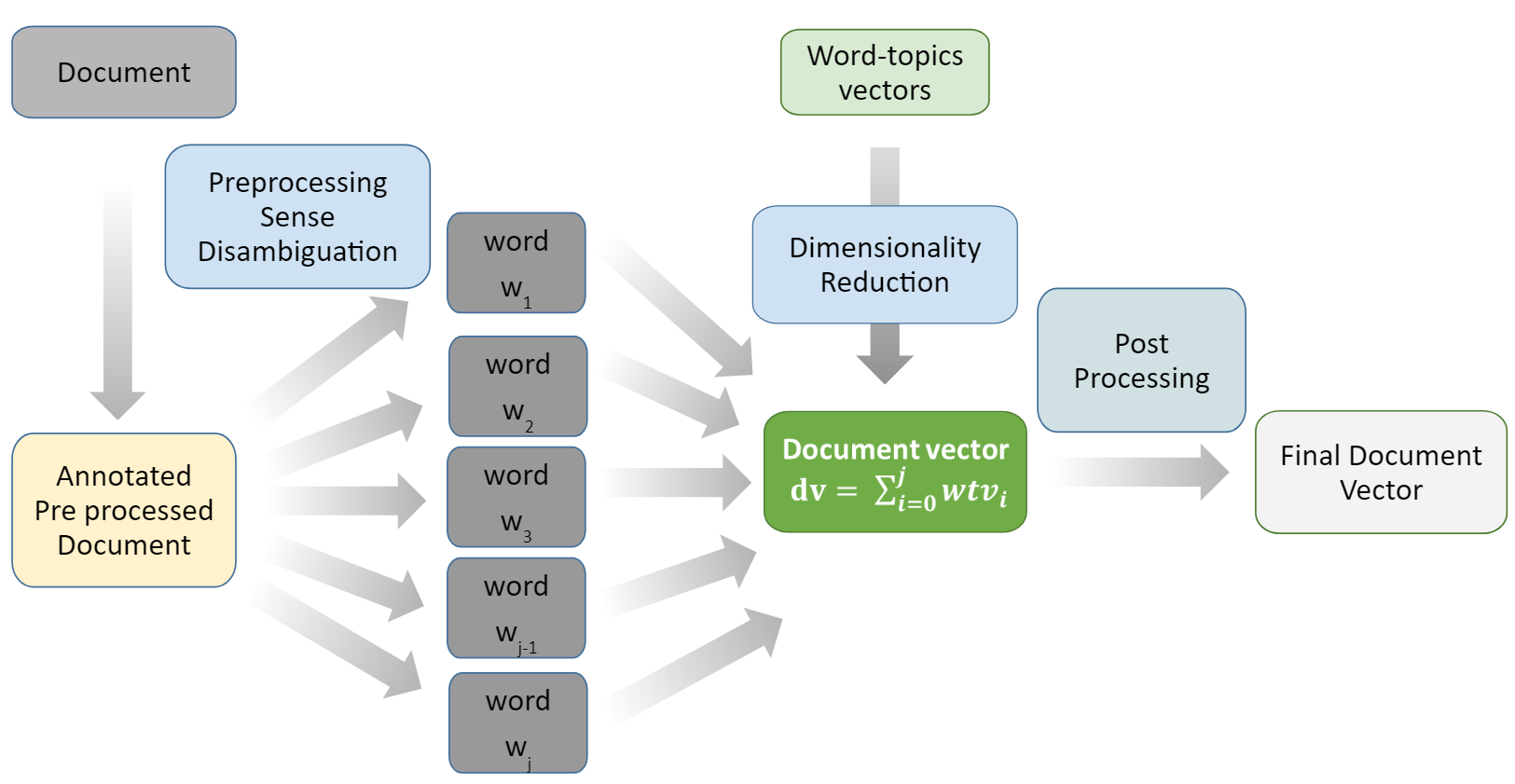}
  \vspace{0.5 em}
  \caption{Flowchart representing final document vector computation.}\label{figure:pipeline2}
\endminipage
% \vspace{-1.5em}
\end{figure*}

\section{Experimental Results}
\label{sec:experiments}

Document embeddings obtained using \method{} can be used as direct features for downstream supervised tasks. We experimented with text classification tasks (see Table \ref{table:expplan}) to show the effectiveness of our embeddings. We evaluated the following questions through our experiments.
\begin{table}
\captionsetup{font=small, skip=0pt}
\vspace{-1.0em}
\caption{Text classification datasets overview.}
\begin{center}
\vspace{-1.0em}
\begin{tabular}{ c c c c c}
\hline
Task & Dataset & \#Classes & \#train / \#test \\ 
\hline
Multi-class & 20NewsGroup & 20 & 11K / 8K\\  
Multi-label & Reuters-21578 & 444 & 13K / 6K \\
\hline
\end{tabular}
\end{center}
\label{table:expplan}
\vspace{-2.4em}
\end{table}

\begin{enumerate}
\item[Q1.] Does disambiguating word-cluster assignments using multi-sense embedding improve classification performance?
\vspace{0.1cm}
\item[Q2.] Does hard thresholding over word-cluster assignments improve performance, space and time complexities? 
\vspace{0.1cm}
\item[Q3.] Is representational noise reduction using Doc2Vec initialization effective? 
\vspace{0.1cm}
\item[Q4.] Can effective lower dimensional manifold be learned from the sparse high dimensional word topic vectors?
\end{enumerate}

\vspace{0.5em}
\noindent \textbf{Baselines: } We considered the following baselines: Bag-of-Words (BoW) \cite{harris54}, Bag of  Word Vector (BoWV) \cite{vivek}, \footnote{\url{https://bit.ly/2X0XfBH}} Sparse Composite Document Vectors (SCDV) \cite{mekala2017scdv}, \footnote{\url{https://bit.ly/36NxGZh}} paragraph vectors \cite{le2014distributed}, pmeans \cite{pmeans}, ELMo \cite{elmo}, Topical word  embeddings (TWE-1) \cite{AAAI159314}, Neural Tensor Skip-Gram Model (NTSG-1 to NTSG-3) \cite{liu2015learning}, tf-idf weighted average word-vector \cite{pranjal2015weighted} and weighted Bag of Concepts (weight-BoC)  \cite{boc}, and BERT \cite{devlin2018bert}. In BoC we built topic-document vectors by counting the member words in each topic. For BERT, we reported the results on the unsupervised pre-trained (pr) model because of a fair comparison to our approach which is also unsupervised.  In Doc2VecC \cite{chen2017efficient} averaging and training the vectors was done jointly with corruption. Also, in SIF \cite{arora2016simple} we used the inverse frequency weights for weighting while averaging word vectors, and finally removed the common components from the average. The results of our proposed embeddings is represented by SCDV-MS in Tables \ref{table:2} and \ref{table:1}. We also compared our representation with topic modeling-based embedding methods, described in the related work.

\vspace{0.5em}
\noindent \textbf{The Experimental Setting: }We learned the word embeddings with Skip-Gram Negative Sampling (SGNS) using commonly used parameters, e.g., negative sample size of $10$, minimum word frequency of $20$, and the window size of $10$. We ensure for usual data cleansing like stop word removal, lemmetization and stemming for all the baselines. In addition, we used simple models such as LinearSVM for multi-class classification and Logistic regression with a OneVsRest setting for the multi-label classification tasks so that we can directly compare our results with the previous approaches which uses the same classifiers. Similar to SCDV, to tune the hyperparameters, we employed a 5-fold cross-validation on the F1 score. We also used the Doc2VecC model \cite{chen2017efficient} to initialize the word embeddings on the annotated corpora for performance improvement. To ensure a fair comparison with SCDV, we fixed the same number of clusters to $60$ and used full covariance for GMM clustering for all experiments based on our best empirical results with cross-validation. We tuned the hard threshold sparsity constant $l$ from range $\{3,5,7\}$ with cross-validation to select the best hyper-parameter for making the word cluster assignments sparse. Moreover, we used AdaGram \cite{bartunov2016breaking} for disambiguating the sense of multi-sense words using a neighborhood of $5$ context words on both sides, so that the window size is $10$. We first ranked our words based on their tf-idf scores; we then selected a practicable number (top $5000$ words) as candidates for the polysemic words. Next, we selected the true polysemic words by applying AdaGram on the candidates. \footnote{\url{https://bit.ly/2Jv6wxX}} The best parameter settings were used to generate baselines results. We used $200$ dimensions for the tf-idf weighted word-vector model, $400$ for the paragraph vector model, $80$ topics and $400$ dimensional vectors for TWE/NTSG/LTSG, and $60$ topics and $200$ dimensional word vectors for SCDV and BOWV. We reported the average of $5$ runs. Our results were robust across multiple runs with a variance of O($10^{-6}$).

\subsection{Text Classification Results} We evaluated the classifier's performance on multi-class classification using several metrics such as accuracy, macro-averaging precision, recall, and macro F1-score. Table \ref{table:1} shows a comparison with multiple state-of-the-art document representations (the first 7 except BERT/ELMo are clustering-based, the next 11 topic-word embeddings based, the next 6 are simple averaging or topic modeling methods) on the 20NewsGroup dataset. We also reported the results (micro F1) on the $20$ classes of 20NewsGroup in Table \ref{table:classwise}. Furthermore, we evaluated the multi-label classification performance using several metrics such as Precision@K, nDCG@k \cite{bhatia2015sparse}, Coverage error, Label ranking average precision score (LRAPS)\footnote{Section $3.3.3.2$ of \url{https://goo.gl/4GrR3M}} and macro F1-score. Table \ref{table:2} shows the results on the Reuters dataset.

\begin{table}
\captionsetup{font=small, skip=0pt}
\caption{Performance on multi-class classification. Values in bold show the best performance using the SCDV-MS embeddings.}
\small
\begin{center}
\setlength\tabcolsep{2.5pt}
\begin{tabular}{ c|c|c|c|c } 
\hline
{\bf Model} & {\bf Accuracy} & {\bf Precision} & {\bf Recall} & {\bf F-measure} \\
\hline
\bf SCDV-MS &\bf 86.19 &\bf 86.20 &\bf 86.18 & \bf 86.16 \\
% \bf SCDV-MS &\bf 86.26 &\bf 86.36 &\bf 86.26 & \bf 86.19 \\
% \bf SCDV(Doc2VecC) &\bf 85.3 &\bf 85.3 &\bf 85.2 &\bf 85.3 \\
% ms-CNN & 86.12 & 85.56 &  85.58 & 85.57\\
\bf R-SCDV-MS &\bf 84.9 &\bf 84.9 &\bf 84.9 &\bf 84.9\\
BERT (pr)\cite{devlin2018bert} & 84.9 & 84.9 & 85.0 & 85.0\\
SCDV \cite{mekala2017scdv} & 84.6 & 84.6 &84.5  & 84.6 \\
RandBin & 83.9 & 83.99 & 83 .9 & 83.76\\
BoWV \cite{vivek} & 81.6 & 81.1 & 81.1 & 80.9\\
pmeans \cite{pmeans} & 81.9 & 81.9 & 81.9 & 81.5\\
Doc2VecC \cite{chen2017efficient}& 84.0 & 84.1 & 84.1 & 84.0\\
BoE \cite{boe} & 83.1 & 83.1 & 83.1 & 83.1 \\
% NTSG-1 & 82.6 & 82.5 & 81.9 & 81.2\\
NTSG-2 \cite{liu2015learning} & 82.5 & 83.7 & 82.8 & 82.4\\
% NTSG-3 & 81.9 & 83.0 & 81.7 & 81.1\\
LTSG \cite{ltsg} & 82.8 & 82.4 & 81.8 & 81.8\\
WTM \cite{wtm}& 80.9 & 80.3 & 80.3 & 80.0\\
ELMo \cite{elmo}& 74.1 & 74.0 & 74.1 & 73.9\\
w2v-LDA \cite{wtvlda} & 77.7 & 77.4 & 77.2 & 76.9\\
TV+MeanWV \cite{tvMeanWV} & 72.2 & 71.8 & 71.5 & 71.6\\
MvTM \cite{mvtm}& 72.2 & 71.8 & 71.5 & 71.6\\
TWE-1 \cite{AAAI159314} & 81.5 & 81.2 & 80.6 & 80.6\\
lda2Vec \cite{lda2vec} & 81.3 & 81.4 & 80.4 & 80.5\\
lda \cite{Blei:2003} & 72.2 & 70.8 & 70.7 & 70.0\\
weight-AvgVec \cite{pranjal2015weighted} &  81.9 & 81.7 & 81.9 & 81.7\\
BoW \cite{pranjal2015weighted} & 79.7 & 79.5 & 79.0 & 79.0\\
weight-BOC \cite{pranjal2015weighted} & 71.8 & 71.3 & 71.8 & 71.4\\
PV-DBoW \cite{le2014distributed}& 75.4 & 74.9 & 74.3 & 74.3 \\
PV-DM \cite{le2014distributed}& 72.4 & 72.1 & 71.5 & 71.5 \\ 
\hline
\end{tabular}
\end{center}
\label{table:1}
\vspace{-0.95em}
\end{table}
\begin{table}
\captionsetup{font=small, skip=0pt}
\caption{Class-wise F1-Score on the 20newsgroup dataset with different document representations.}
\small
\begin{center}
\setlength\tabcolsep{2.5pt}
\begin{tabular}{ c|c|c|c }
\hline
{\bf Class Name} & {\bf SCDV} & {\bf SCDV-MS} & {\bf R-SCDV-MS}\\
\hline
alt.atheism & 80.14 &\bf 81.35 & 80.39 \\
comp.graphics &\bf 78.99 & 76.84 & 76.95 \\
comp.os.ms-windows.misc & 75.65 & 77.65 &\bf 78.28 \\
comp.sys.ibm.pc.hardware  & 72.08 &\bf 73.43 & 68.38 \\
comp.sys.mac.hardware & 82.15 &\bf 86.82 & 80.16 \\
comp.windows.x & 81.8 & 82.97 &\bf 83.27 \\
misc.forsale & 82.8 &\bf 85.13 & 84.99 \\
rec.autos & 89.06 &\bf 92.53 &  91.77 \\
rec.motorcycles & 94.27 &\bf 96.11 & 94.27 \\
rec.sport.baseball & 93.57 &\bf 96.47 & 93.68 \\
rec.sport.hockey &\bf 97.27 & 96.78 & 96.41 \\
sci.crypt & 93.1 & 92.82 &\bf 93.5 \\
sci.electronics & 77.38 &\bf 77.45 & 74.25 \\
sci.med & 88.58 &\bf 92.30 & 91.57 \\
sci.space & 90.33 &\bf 91.40 & 90.71 \\
soc.religion.christian & 89.56 &\bf 89.97 & 89.76 \\
talk.politics.guns & 80.69 &\bf 84.18 & 83.05 \\
talk.politics.mideast & 95.96 & 95.95 &\bf 96.1 \\
talk.politics.misc & 69.33 & 73.49 &\bf 73.67 \\
talk.religion.misc & 65.53 &\bf 65.54 & 60.48 \\
\hline
\end{tabular}
\end{center}
\label{table:classwise}
\end{table}

\begin{table}
\captionsetup{font=small, skip=0pt}
\small
\begin{center}
\caption{Performance on various metrics for multi-label classification on the Reuters dataset. Values in bold show the best performance using the SCDV-MS algorithm.}
\vspace{1.0em}
\label{table:2}
\setlength\tabcolsep{1.5pt}
\begin{tabular}{ c|c|c|c|c|c|c } 
 \hline
{\bf Model} & \multicolumn{1}{p{1.0cm}|}{\centering {\bf Prec} \\ {\bf @1}} & \multicolumn{1}{p{1.0cm}|}{\centering {\bf Prec} \\ {\bf @5}} & \multicolumn{1}{p{1.0cm}|}{\centering {\bf nDCG} \\ {\bf @5}} & \multicolumn{1}{p{1.0cm}|}{\centering {\bf Cover.} \\ {\bf Error}} & {\bf LRAPS} & \multicolumn{1}{p{1.0cm}}{\centering {\bf F1}\\ {\bf Score}}\\
 \hline
 \bf SCDV-MS & \bf 95.06 & \bf 37.56 & \bf 50.20 & \bf 5.87 & \bf 94.21 & \bf 82.71 \\
% \bf SCDV-MS & \bf 94.05 & \bf 37.14 & \bf 49.67 & \bf 6.40 & \bf 93.42 & \bf 82.37 \\
% \bf SCDV(Doc2VecC)    &\bf 93.88    &\bf 37.13    &\bf 49.62    &\bf 6.55 &\bf 93.22 &\bf 82.01 \\
\bf R-SCDV-MS &\bf 93.56 &\bf 37.00 &\bf 49.47 &\bf 6.74 &\bf 92.96 &\bf 81.94 \\
BERT (pr) & 93.8 & 37 & 49.6 & 6.3 & 93.1 & 81.9\\
SCDV & 94.00 & 37.05 & 49.6 & 6.65    & 93.34 & 81.77 \\
Doc2VecC & 93.45 & 36.86 & 49.28 & 6.83 & 92.66 & 81.29\\
pmeans & 93.29 & 36.65 & 48.95 & 7.66 & 91.72 & 77.81\\
BoWV & 92.90 & 36.14 & 48.55 & 8.16 & 91.46 & 79.16 \\
TWE-1  & 90.91 & 35.49 & 47.54 & 8.16 & 91.46 & 79.16\\
PV-DM  & 87.54 & 33.24 & 44.21 &  13.15 & 86.21 & 70.24\\ 
PV-DBoW  & 88.78 &  34.51& 46.42 & 11.28 & 87.43 & 73.68\\
% AvgVec  & 89.09 & 34.73 & 46.48 & 9.67 & 87.28 & 71.91\\
tfidf AvgVec & 89.33 & 35.04 & 46.83 & 9.42 & 87.90 & 71.97\\
\hline
\end{tabular}
\end{center}
\vspace{-1.25em}
\end{table}
\noindent \textbf{Datasets: }We evaluated our approach by running multi-class classification experiments on the 20NewsGroup dataset, \footnote{\url{http://qwone.com/\~jason/20Newsgroups/}} and multi-label classification experiments on the Reuters-21578 dataset. \footnote{\url{https://goo.gl/NrOfu}} For more details on dataset statistics refer to Table \ref{table:expplan}. We used \textit{script} for datasets preprocessing. \footnote{\url{https://bit.ly/2PXDdXj}} 

\begin{table}
\centering
\captionsetup{font=small, skip=0.0pt}
\caption{Ablation Study reporting F1 scores. In $\pm x$, $x$ is the variance across several runs.}
\vspace{1.0em}
\small
\begin{tabular}{c|c|c}
\hline
\bf Ablation (w/o)  &\bf 20NewsGroup &\bf Reuters\\ \hline
Sparsity & 85.28 $\pm$ 0.002 & 82.17 $\pm$ 0.001\\
Doc2VecC & 85.41 $\pm$ 0.001 & 82.08 $\pm$ 0.002\\
MultiSense & 85.16 $\pm$ 0.001 & 82.43 $\pm$ 0.001\\
All & 84.61 $\pm$ 0.004 & 81.77 $\pm$ 0.003 \\
None & \bf 86.16 $\pm$ 0.002 &\bf 82.71 $\pm$ 0.002\\ \hline
\end{tabular}
\label{tab:ablationStudy}
\end{table}
\begin{table}
\captionsetup{font=small, skip=0pt}
\caption{Performance in terms of accuracy with various dimensionality reduction methods on the 20NewsGroup dataset. Similar results were acquired for precision, recall, and F1 score.}
\small
\begin{center}
\begin{tabular}{ c|c|c|c }
\hline
\bf Dimension & \multicolumn{1}{p{1.5cm}|}{\centering {\bf Random} \\ {\bf Projection}} & \multicolumn{1}{p{1.5cm}|}{\centering {\bf PCA} \\ {\bf (SubSpace)}} & \bf Autoencoder \\
\hline
200 & 78.97 & 80.62 &  81.44 \\
500 & 82.19 & 83.14 & 83.83 \\
1000 & 83.75 & 83.80 & 84.31 \\
2000 & 84.47 & 84.34 & 84.80 \\
3000 & \bf 84.94 &\bf 84.86 &\bf 84.90 \\
\hline
\end{tabular}
\end{center}
\label{table:DimRed20NewsGroup}
\end{table}

\begin{figure*}[!htb]
\minipage{0.33\textwidth}
\captionsetup{font=small, skip=0pt}
  \includegraphics[width=\linewidth]{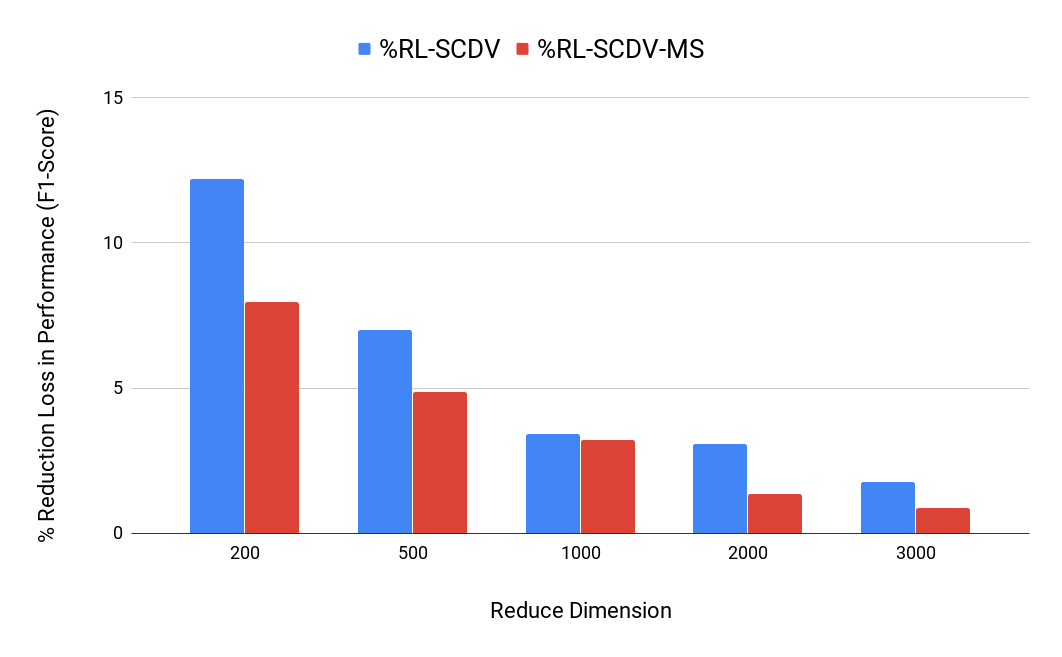}
   \vspace{-1.2em}
  \caption{Percentage loss in F1-Score (\%RL) after Random Projection-based  $\vec{wtv}$ dimensionality reduction on 20NewsGroup.}\label{fig:randProj}
\endminipage\hfill
\minipage{0.33\textwidth}
\captionsetup{font=small, skip=0pt}
  \includegraphics[width=\linewidth]{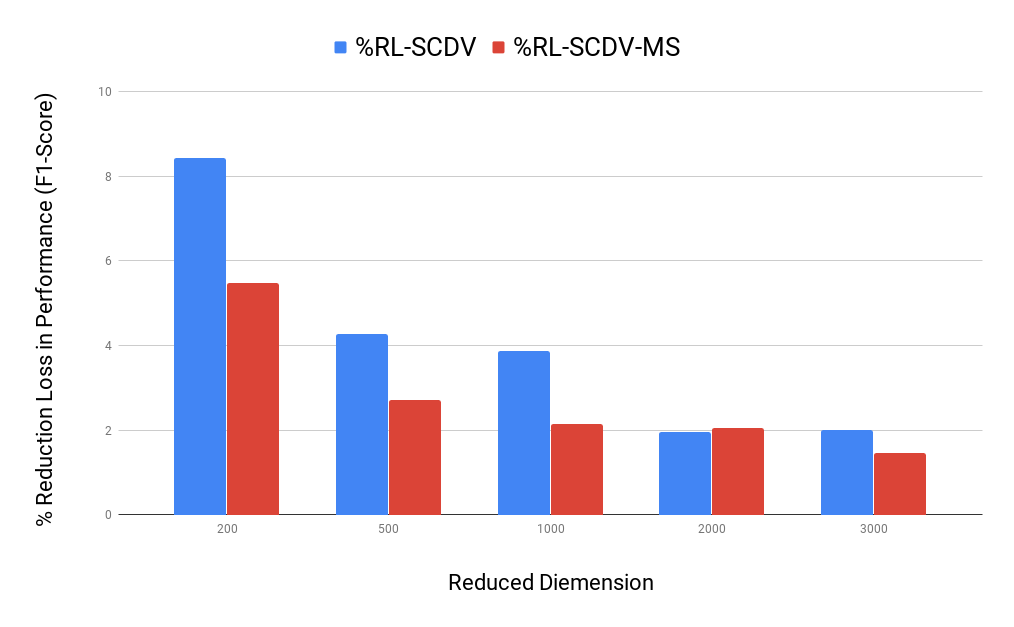}
   \vspace{-1.2em}
  \caption{Percentage loss in F1-Score (\%RL) after Autoencoder-based  $\vec{wtv}$ dimensionality reduction on 20NewsGroup.}\label{fig:autoProj}
\endminipage\hfill
\minipage{0.31\textwidth}%
\captionsetup{font=small, skip=0pt}
  \includegraphics[width=\linewidth]{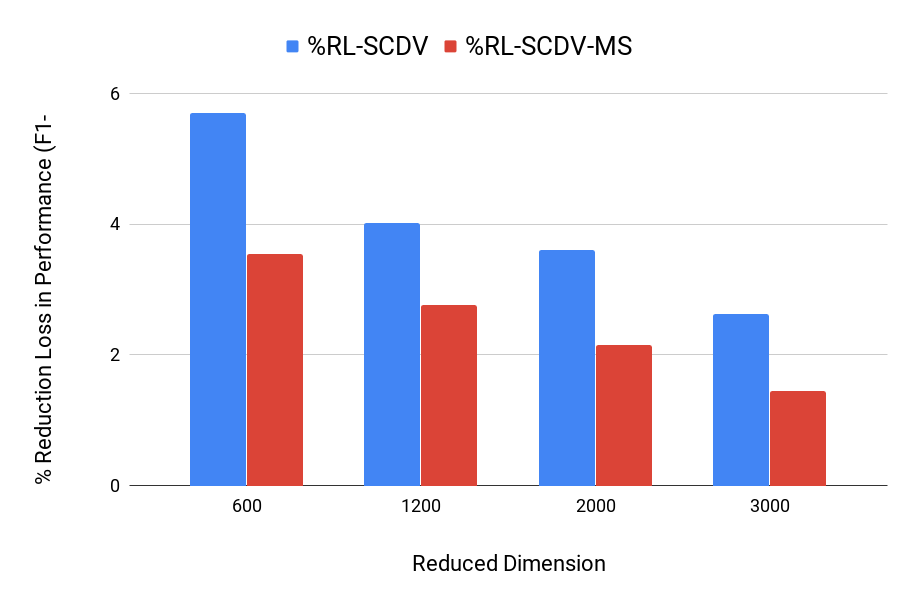}
  \vspace{-1.2em}
  \caption{Percentage loss in F1-Score (\%RL) after PCA (Subspace)-based $\vec{wtv}$ dimensionality reduction on 20NewsGroup.}\label{fig:pcaProj}
\endminipage
\vspace{-1.0em}
\end{figure*}

\begin{table*}
\captionsetup{font=small, skip=0pt}
\caption{Performance of Convolutional Neural Network (CNN) for multi-class text classification on 20NewsGroup with the original word embeddings ($200$ dimensions) and the reduced word topic vectors ($2000$ dimensions). In $\pm x$, $x$ is the variance across several runs.}
\small
\begin{center}
\begin{tabular}{ c|c|c|c|c|c }
\hline
\bf Embedding &\bf Dimen &\bf Accuracy &\bf Precision &\bf Recall &\bf F1-Score \\
\hline
Word2Vec & 200 & 82.07 $\pm$ 0.003  & 82.22 $\pm$ 0.005 & 81.9 $\pm$ 0.004 & 81.9 $\pm$ 0.005\\
Reduce Word Topic Vector & 200 &\bf 82.13 $\pm$  0.002 &\bf 82.36 $\pm$ 0.003 &\bf 82.06 + $\pm$ 0.003 &\bf 82.05 $\pm$ 0.003\\ \hline
Word2Vec & 2000 & 82.31 $\pm$ 0.004 & 82.87 $\pm$ 0.005 &   82.31 $\pm$ 0.004 & 82.38 $\pm$ 0.005 \\
Reduce Word Topic Vector & 2000 &\bf 82.85 $\pm$  0.001 &\bf 83.18 $\pm$ 0.004  &\bf 82.64 $\pm$ 0.002 &\bf 82.68 $\pm$ 0.003 \\
\hline
\end{tabular}
\end{center}
\label{table:CNNResults20NewsGroup}
\vspace{-0.5em}
\end{table*}
\begin{table*}[!ht]
\vspace{-0.50em}
\captionsetup{font=small, skip=0pt}
\caption{Time and Space complexity analysis of embedding methods. Bold values represent the best results.}
\small
\setlength\tabcolsep{3.2pt}
\begin{center}
\begin{tabular}{ c|c|c|c|c|c|c|c|c|c|c }
\hline
\bf Method  &\bf Vocab  & \multicolumn{1}{p{0.8cm}|}{\centering {\bf $\vec{wtv}$ } \\ {\bf Dim}}  & \multicolumn{1}{p{1.5cm}|}{\centering{\bf$\vec{wtv}$ } \\ {\bf Sparsity($\%$)}} & \multicolumn{1}{p{1.5cm}|}{\centering {\bf $\vec{dv}$ } \\ {\bf Sparsity($\%$)}} & \multicolumn{1}{p{1.3cm}|}{\centering {\bf Cluster} \\ {\bf (sec)}} & \multicolumn{1}{p{1.3cm}|}{\centering {\bf Feature} \\ {\bf ($\mu$ sec)}} & \multicolumn{1}{p{0.9cm}|}{\centering \bf {Predict} \\ {($\mu$ sec)}} & \multicolumn{1}{p{1.3cm}|}{\centering \bf {Training}\\{(min)}} & \multicolumn{1}{p{1.3cm}|}{\centering \bf {Model}\\{Size  (KB)}} & \multicolumn{1}{p{1.3cm}}{\centering \bf {$\vec{wtv}$ }\\{Space(MB)}} \\
\hline
SCDV  & 15591 & 12000 & 1 &\bf 81 &\bf 242 & 2.56 & 119 & 82 & 1900 & 748\\
% \hline
SCDV-MS & 25466 & 12000 &\bf 98 & 74 & 569 &\bf 0.06 & 111 & 79 & 1900 &\bf 71\\
% \hline
R-SCDV-MS & 25466 & 2000 & 0 & 0 & 576 & 0.86 &\bf 14 &\bf 66 & \bf333 & 203\\
\hline
\end{tabular}
\end{center}
\label{table:timespacecomplex}
\vspace{-1.3em}
\end{table*}

\begin{table}
\captionsetup{font=small, skip=0pt}
\caption{PCA based subspace rank reduction criteria.}
\small
\begin{center}
\begin{tabular}{ c|c }
\hline
\bf Red-Dim & \bf Subspace Rank Reduction Criteria \\
\hline
$500$ & rank $>$ $10$ reduce to $10$ else the original rank\\
$1000$ & rank $>$ $20$ reduce to $15$ else the original rank\\
$2000$ & rank $>$ $100$ reduce to $30$ else the original rank\\
$3000$ & rank $>$ $100$ reduce to $50$ else the original rank\\
\hline
\end{tabular}
\end{center}
\label{table:pcarankreduction}
\vspace{-1.5em}
\end{table}

\begin{table}
\captionsetup{font=small, skip=0pt}
\small
\begin{center}
% \vspace{-0.3em}
\caption{Performance on reduced word topic vectors using several reduction techniques on various dimensions for multi-label classification -- the Reuters dataset.}
\vspace{1.0em}
\label{table:ReduceReuters}
\setlength\tabcolsep{1.5pt}
\begin{tabular}{ c|c|c|c|c|c|c|c } 
 \hline
% {\bf Method} & {\bf Dimen} & \multicolumn{1}{p{1.5cm}|}{\centering {\bf Prec@1} \\ {\bf nDCG@1}} & {\bf Prec@5} & {\bf nDCG@5} & \multicolumn{1}{p{1.3cm}|}{\centering {\bf Cover} \\ {\bf Error}} & {\bf LRAP} & \multicolumn{1}{p{1.5cm}}{\centering {\bf F1}\\ {\bf Score}}\\
\bf Method & \bf Dimen & \bf Prec@1 & \bf Prec & \bf nDCG &\bf Cover & \bf LRAP & \bf F1\\
 & &\bf nDCG &\bf @5 & @5 &\bf Error & &\bf Score \\
 \hline
& 500 & 92.14 & 36.53 & 48.74 & 8.02 & 91.34 & 79.96 \\
Auto & 1000 & 92.95 & 36.82 & 49.17 & 7.14 & 92.32 & 81.05 \\
Encoder & 2000 & 93.39 & 36.95 & 49.39 & 6.87 & 92.75 & 81.65 \\
& 3000 &\bf 93.56 &\bf 37 &\bf 49.47 &\bf 6.74 &\bf 92.96 &\bf 81.94 \\
\hline
& 500 & 91.98 & 36.26 & 48.47 & 7.41 & 91 & 79.03 \\
Random & 1000 & 92.59 & 36.62 & 48.91 & 6.98 & 91.84 & 80.39 \\
Projection & 2000 &\bf 93.39 & 36.83 & 49.26 & 6.95 & 92.59 & 81.12 \\
& 3000 & 93.32 &\bf 36.91 &\bf 49.33 &\bf 6.78 &\bf 92.75 &\bf 81.39 \\
\hline
& 500 & 90.73 & 36.03 & 48.06 & 8.40 & 90.11 & 78.06 \\
PCA & 1000 & 92.15 & 36.55 & 48.78 & 7.44 & 91.48 & 80 \\
(SubSpace) & 2000 & 92.95 & 36.87 & 49.22 & 6.86 & 92.30 & 81.2 \\
& 3000 &\bf 93.38 &\bf 36.97 &\bf 49.4 &\bf 6.69 &\bf 92.8 &\bf 81.48 \\
\hline
\end{tabular}
\end{center}
\end{table}
\begin{table}
\captionsetup{font=small, skip=0pt}
\vspace{-0.5em}
\caption{Time complexity for dimensionality reduction of word topic vectors to a $2000$-dimension dense representation using various reduction techniques on 20NewsGroup}
\small
\begin{center}
\begin{tabular}{ c|c|c|c }
\hline
\bf Method & \multicolumn{1}{p{1.5cm}|}{\centering {\bf Random} \\ {\bf Projection}} & \multicolumn{1}{p{1.5cm}|}{\centering {\bf PCA} \\ {\bf (SubSpace)}} & \bf Autoencoder \\
\hline
Time (sec)  & 35 & 66 & 608\\
\hline
\end{tabular}
\end{center}
\label{table:reductimecomplex}
\vspace{-1.5em}
\end{table}
\vspace{0.5em}
\noindent \textbf{Results and Analysis.} We observed that our modified embeddings (SCDV-MS) with Doc2VecC-initialized word vectors, direct thresholding on word cluster assignments, and multi-sense disambiguation using AdaGram outperforms all earlier embeddings on both the 20NewsGroup and the Reuters datasets. From class-wise results in Table \ref{table:classwise}, we notice a consistent performance improvement where we are outperforming SCDV in 18 out of 20 classes. It should be noted that the improvement on Reuters is not as great as the 20NewsGroup dataset due to the fact that the number of unique polysemic words in Reuters ($250$) is significantly fewer than 20NewsGroup ($1000$); thus each word is assigned to only one cluster. Therefore, the use of AdaGram for sense disambiguation and the sparsity operation over the word-cluster assignments does not improve the performance by a large margin. We verified this claim in the ablation study below. We can conclude that our modifications yield notable improvements if the dataset has more multi-sense words.

\vspace{0.5em}
\noindent \textbf{Ablation Study: } To understand the contributions of each of the three modifications, we compared five different versions of our embeddings. In the first version, we ablated the sparsity of the word-cluster assignments and applied sparsity directly on the document vectors similar to SCDV while keeping the Doc2VecC multi-sense word embeddings and the sense annotated corpus intact. In the second version, we ablated the Doc2VecC embeddings with normal SGNS embedding while keeping the word topic vector sparsity, and the sense annotated corpus intact. In the third version, we ablated multi-sense embeddings and the annotation of the corpus while keeping the Doc2VecC word training and the word topic vector sparsity intact. We also compared our results with an all ablation approach, i.e., the SCDV baseline and a none ablation approach, i.e., our new embeddings in SCDV-MS. Table \ref{tab:ablationStudy} shows the results obtained with ablation on 20NewsGroup and Reuters datasets. We obtained the best performance with the none ablation approach, i.e., SCDV-MS. Thus, we can conclude that all three modifications is needed to yield the best performance. Multi-sense is the most pivotal improvement for 20newsgroup since by ablating it we observed the lowest performance out of ablating each of the three modifications. Whereas on the Reuters dataset, multi-sense was the least important because of fewer multi-sense words. On Reuters, the noise removal at word level representations was the most important.

\vspace{0.5em}
\noindent \textbf{Comparison with Contextual Embeddings: } SCDV-MS is a lot simpler than unsupervised contextual embeddings like BERT (pr) and ELMo, but it outperformed them. We presume that SCDV-MS's concentration on capturing semantics (local and global) in sparse high dimension representations instead of capturing both semantics and syntax in single lower dimensional continuous representations (what BERT does) is the reason behind our method's superior performance. Because understanding syntax is not as influential as semantics in our classification and similarity tasks.

\subsection{wtv's Lower Dimensional Manifold Learning}
\label{subsec:dimreduction}
\begin{figure}
\vspace{-0.5em}
\captionsetup{font=small, skip=0.0pt}
\centering
\includegraphics[width=0.30\textwidth]{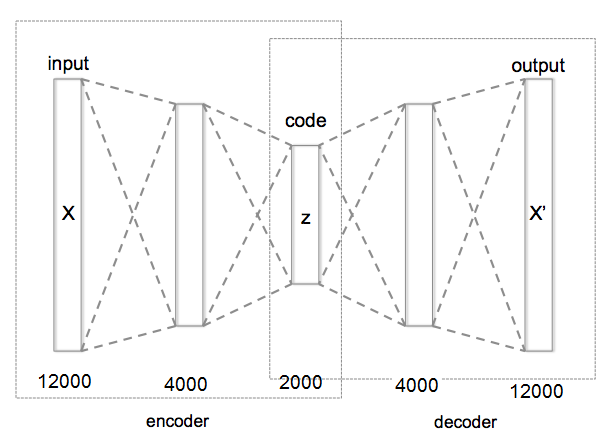}
\vspace{1.0em}
\caption{The AutoEncoder architecture ($12000-4000-2000-4000-12000$) used to reduce the word topic vectors from $12000$ dimensions to $2000$ dimensions. }
\label{fig:ourautoarch}
\vspace{-1.5em}
\end{figure}

We tried three popular lower-dimensional manifold learning algorithms to reduce high dimensional sparse word topic vectors to lower-dimensional word embeddings, namely Random Projection \cite{achlioptas2003database}, PCA (Subspace) \cite{abdi2010principal}, and Autoencoders  \cite{vincent2010stacked}. For PCA (Subspace), we observed that not all subspaces ($200$ dimensions) of word topic vectors have a complete full rank ($200$). Most of the subspaces were of ranks much smaller than $200$ (rank $\leq$ $200$). Therefore, we applied PCA on each of $200$ dimension subspaces separately and concatenated the subspace reduced vectors. We refer to Table \ref{table:pcarankreduction} as the criteria for PCA-based subspace rank reduction. For Auto-encoders, we used the standard architecture for reducing the word topic vectors from $12000$ to $2000$ dimensions, through a intermediate layer of $4000$ dimensions (see Figure \ref{fig:ourautoarch}). We used the mean squared error minimization, $\tanh$ non-linear activation, and Adam optimization routine for training the autoencoders. We used the initial learning rate of $0.001$, $\beta$$_1=0.9$ and $\beta$$_2=0.999$ for Adam optimization routine. 

\vspace{0.2cm}
\noindent \textbf{Results:} Table \ref{table:DimRed20NewsGroup} shows the performance of the dimensionality reduction techniques such as Random Projection, PCA (Subspace), and Autoencoders on the 20NewsGroup dataset. We observed that autoencoders outperform other reduction methods because they easily fit any non-linear function. Also, we observed only a small decrease of $~1\%$ in the performance after a reduction to $2000$ dimensions with most methods. This decrease is associated to loss due to data compression and can be explained by information theory. However, word topic vectors can be efficiently projected into a lower-dimensional manifold without a significant loss in the performance. 
We compared the percentage loss in performance (F1-Score) \%RL = $\frac{\text{Orignal} - \text{Reduced}}{\text{Orignal}}$ on text classification with a dimension-reduced $\vec{wtv}$ through random projection for both SCDV and SCDV-MS on 20newsgroup. In Figures \ref{fig:randProj}, \ref{fig:autoProj} and \ref{fig:pcaProj} we observe that the $\%RL$ loss in SCDV-MS's F1-Score is distinctively less compared to SCDV for all reduction methods. Furthermore, we observed that the reduction time for SCDV-MS was shorter than SCDV, particularly for random projection, because SCDV-MS $\vec{wtv}$ are sparser than SCDV $\vec{wtv}$. We also tried a direct reduction of final document representations which yielded a poor performance and took a longer reduction time for both embeddings. Overall, reducing SCDV-MS $\vec{wtv}$ is much more effective than reducing SCDV $\vec{wtv}$ or document vectors. Similar observations for a multi-label classification task on the Reuters dataset, see Table \ref{table:ReduceReuters}. 

% \begin{figure}[!ht]
% \vspace{-1.20em}
% \captionsetup{font=small, skip=0pt}
% \centering
% \includegraphics[width=0.35\textwidth]{random-projection.png}
% \caption{Percentage loss in F1-Score (\%RL) after Random Projection-based  $\vec{wtv}$ dimensionality reduction on 20NewsGroup.}
% \label{fig:randProj}
% \vspace{-1.20em}
% \end{figure}
% \begin{figure}[!ht]
% \captionsetup{font=small, skip=0pt}
% \centering
% \includegraphics[width=0.32\textwidth]{red_autoencoder.png}
% \caption{Percentage loss in F1-Score (\%RL) after Autoencoder-based  $\vec{wtv}$ dimensionality reduction on 20NewsGroup.}
% \label{fig:autoProj}
% \vspace{-0.95em}
% \end{figure}
% \begin{figure}[!ht]
% \captionsetup{font=small, skip=0pt}
% \centering
% \includegraphics[width=0.32\textwidth]{pca-subspace.png}
% \caption{Percentage loss in F1-Score (\%RL) after PCA (Subspace)-based $\vec{wtv}$ dimensionality reduction on 20NewsGroup.}
% \label{fig:pcaProj}
% \vspace{-1.0em}
% \end{figure}

\vspace{0.5em}
\noindent \textbf{Application to Deep Learning: }One significance of our reduction of $\vec{wtv}$ is that they can be used as direct word embeddings in popular deep learning architectures such as CNNs on downstream classification tasks. We used the same architecture (see the supplementary material, Figure 8 \footnote{ \url{https://bit.ly/33473wk}}) for both embeddings. Employing CNN, our results outperformed the original word embeddings of the same dimension for the 20NewsGroup classification task, shown in Table \ref{table:CNNResults20NewsGroup}.
% \begin{table}[!htbp]
% \captionsetup{font=small, skip=0pt}
% \caption{Time complexity for dimensionality reduction of word topic vectors to a $2000$-dimension dense representation using various reduction techniques on 20NewsGroup }
% % \vspace{0.1cm}
% \small
% \begin{center}
% \begin{tabular}{ c|c|c|c }
% \hline
% \bf Method & \multicolumn{1}{p{1.5cm}|}{\centering {\bf Random} \\ {\bf Projection}} & \multicolumn{1}{p{1.5cm}|}{\centering {\bf PCA} \\ {\bf (SubSpace)}} & \bf Autoencoder \\
% \hline
% Time (sec)  & 35 & 66 & 608\\
% \hline
% \end{tabular}
% \end{center}
% \label{table:reductimecomplex}
% \vspace{-0.5cm}
% \end{table}
\subsection{Time and Space Complexity}
\label{subsec:timespacecomplexity}
Table \ref{table:timespacecomplex} illustrates empirical results for time and space complexities on SCDV ($12000$ dimensions), SCDV-MS ($12000$ dimensions) and reduced R-SCDV-MS ($2000$ dimensions). 

\vspace{0.3em}
\noindent \textbf{Feature Formation Time}: Due to the direct thresholding of word cluster assignments in SCDV-MS, the word topic vectors ($\vec{wtv}$) are extremely sparse. SCDV-MS $\vec{wtv}$ has only $2\%$ of active attributes ($98\%$ sparse), whereas SCDV $\vec{wtv}$ has $99\%$ of active attributes (only $1\%$ sparse). Therefore, we can use an efficient sparse operation (sparse addition and multiplication) over sparse vectors to speedup feature formation. We observed that by adding sparsity we can reduce the feature formation time by a significant factor of $43$.

\vspace{0.3em}
\noindent \textbf{Overall Prediction Time}: Overall, due to enhanced feature formation and reduced $\vec{wtv}$ loading time, SCDV-MS will predict faster compared to the original SCDV. However, we observed an insignificant difference in the prediction time and model size as SCDV sparsifies the final document vectors (both equally sparse). Furthermore, after reducing the SCDV-MS $\vec{wtv}$ to $2000$-dense dimension features using auto-encoders, we observed a distinctive reduction of $8.5$ times in the prediction time. One can directly store the reduced $\vec{wtv}$ for the complete vocabulary instead of the reduction model.  Refer to Table \ref{table:reductimecomplex} for SCDV-MS $\vec{wtv}$ reduction timing. Furthermore, one can directly also reduce the words appearing in the documents i.e. use a real-time reduction model during prediction.
%  This approach can be faster than reducing complete vocabulary all at once.

\vspace{0.3em}
\noindent \textbf{Vector Space Complexity, Training Time, and Model Size}:  We only require $0.1$ of the original space to store sparse $\vec{wtv}$. We achieved these improvements despite having $1.63$ times of the size of the original vocabulary due to multi-sense word embeddings. The projected $\vec{wtv}$ is $3$ times larger than the $\vec{wtv}$ of SCDV-MS; however, it is $3.7$ times smaller than the $\vec{wtv}$ in SCDV. SCDV is marginally ($1.1 $ times) sparser than SCDV-MS due to manual thresholding of document vectors. SCDV-MS also has a slightly faster training time compared to the original SCDV. We also observed that our training model on reduced vectors (R-SCDV-MS) is $6$ times smaller than SCDV, and the training process of our classifier is $1.25$ times faster than SCDV. 

\section{Conclusion}
\label{sec:conclusions}
In this paper, we proposed several novel modifications to overcome the shortcomings of SCDV, a  state-of-the-art document embedding method. Our proposed \method{}, outperformed the previous embedding (including SCDV and BERT (pr)) on the downstream tasks of text classification. Overall, we have shown that disambiguating multi-sense words based on context words (adjacent words) can lead to better document representations. Sparsity in representation is helpful for effective and efficient lower-dimensional manifold representation learning. Representation noise in words level can have a significant impact on the downstream tasks. 

% \ack We would like to thank the referees for their comments, which
% helped improve this paper considerably

\bibliography{reference}

\end{document}